\documentclass[conference,12pt,onecolumn,draftclsnofoot]{IEEEtran}
\IEEEoverridecommandlockouts
\usepackage[a4paper]{geometry}
\usepackage{cite}
\usepackage{amsmath,amssymb,amsfonts}
\usepackage{graphicx}
\usepackage{textcomp}
\usepackage{xcolor}
\usepackage{svg}
\usepackage{float}

\usepackage{titlesec}
\renewcommand{\thesection}{\arabic{section}} % Arabic numbering for sections (e.g., 1, 2, 3)
\renewcommand{\thesubsection}{\arabic{section}.\arabic{subsection}} % Arabic numbering for subsections (e.g., 1.1, 1.2)
\renewcommand{\thesubsubsection}{\arabic{section}.\arabic{subsection}.\arabic{subsubsection}} % Arabic numbering for subsubsections (e.g., 1.1.1)

\titleformat{\section}%
  [hang] % Hanging indent format
  {\normalfont\bfseries\Large} % Font and size
  {\thesection.} % Section label (adds the number before the title)
  {1em} % Space between the label and the title
  {} % Before code (empty in this case)

\titleformat{\subsection}%
  [hang] % Hanging indent format
  {\normalfont\bfseries\large} % Font and size for subsections
  {\thesubsection.} % Subsection label (adds the number before the title)
  {1em} % Space between the label and the title
  {} % Before code (empty in this case)

\titleformat{\subsubsection}%
  [hang] % Hanging indent format
  {\normalfont\bfseries} % Font and size for subsubsections
  {\thesubsubsection.} % Subsubsection label (adds the number before the title)
  {1em} % Space between the label and the title
  {} % Before code (empty in this case)

\usepackage{tocloft} % Customize table of contents
\usepackage{hyperref} % Enable clickable links

\hypersetup{
    colorlinks=true,
    linkcolor=black, % Color for internal links
    urlcolor=black,  % Color for external links
    citecolor=blue  % Color for citation links
}

\def\BibTeX{{\rm B\kern-.05em{\sc i\kern-.025em b}\kern-.08em\TeX}}

\emergencystretch=\maxdimen
\title{Ethics and Technical Aspects of Generative AI Models in Digital Content Creation}

\author{\IEEEauthorblockN{Atahan Karagöz}
\IEEEauthorblockA{\textit{Department of Computer Science} \\
\textit{University of Basel}\\
Basel, Switzerland \\
atahan.karagoez@stud.unibas.ch}
}

\begin{document}

\maketitle
\thispagestyle{empty} % No page number on title/abstract page

\begin{abstract}
\noindent Generative AI models like GPT-4o and DALL-E 3 are reshaping digital content creation, offering industries tools to generate diverse and sophisticated text and images with remarkable creativity and efficiency. This paper examines both the capabilities and challenges of these models within creative workflows. While they deliver high performance in generating content with creativity, diversity, and technical precision, they also raise significant ethical concerns. Our study addresses two key research questions: (a) how these models perform in terms of creativity, diversity, accuracy, and computational efficiency, and (b) the ethical risks they present, particularly concerning bias, authenticity, and potential misuse. Through a structured series of experiments, we analyze their technical performance and assess the ethical implications of their outputs, revealing that although generative models enhance creative processes, they often reflect biases from their training data and carry ethical vulnerabilities that require careful oversight. This research proposes ethical guidelines to support responsible AI integration into industry practices, fostering a balance between innovation and ethical integrity.
\end{abstract}

% Start a new page, reset page numbering, and begin Table of Contents
\newpage
\pagenumbering{arabic} % Start page numbers in Arabic numerals from this page onward
\setcounter{page}{2} % Start numbering from 2

\setlength{\parskip}{0pt}
\tableofcontents
\newpage
\setlength{\parskip}{0pt}

\section{Introduction}
Generative AI transformed the world of digital content creation, as it made computers generate text, images, audio, and even video that closely resemble human-made content. Applications range from automating writing tasks via tools like GPT and DALL-E to creating unique visual arts, up to realistic media, with major implications for industries like advertising, entertainment, and journalism. This also raises ethical questions concerning responsible use in the areas of authenticity, intellectual property, and biases \cite{barocas2020fairness}.

This paper examines the dual aspects of generative AI in content creation: technical potential and ethical impact \cite{bommasani2022opportunities}. Its objectives include analyzing model performance, assessing ethical risks, and promoting responsible practices. Through technical and ethical experiments, we aim to highlight generative AI's capabilities and limitations, contributing to the dialogue on balancing innovation with ethical responsibility in AI-driven content creation.

\section{Literature Review}
\subsection{Generative AI Technical Landscape}
Generative AI, using advanced neural networks, enables machines to create creative works across media. Transformer-based systems like GPT and DALL-E are now essential in AI-driven content creation \cite{brown2020scaling}. GPT models generate realistic conversational text, while DALL-E creates diverse, complex images from text prompts. However, training these models requires vast data and computational resources, making them powerful yet costly. Their high resource demands also raise concerns about environmental sustainability and accessibility, potentially limiting smaller organizations' adoption.

\begin{figure}[H]
  \centering
  
  \includegraphics[width=\columnwidth]{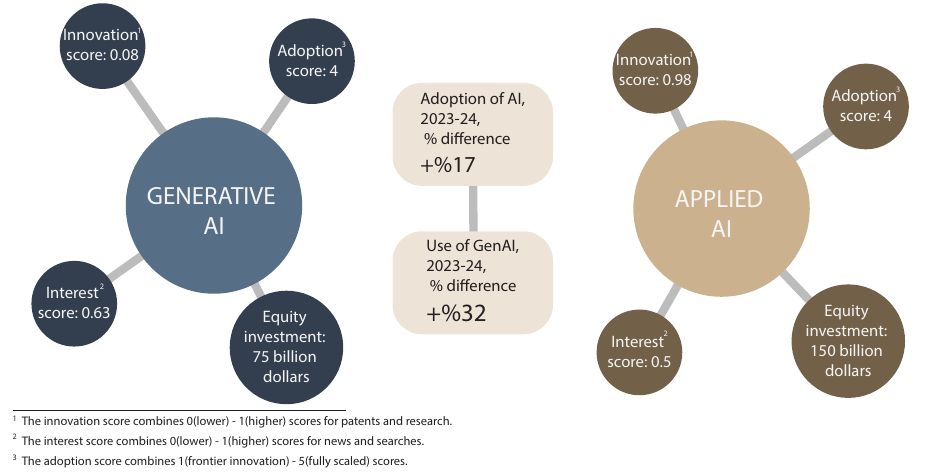}
  \caption{Adoption trends of Generative AI and Applied AI, showing innovation, adoption, interest, and investment. (Data Source: \cite{mckinsey2024ai}\cite{mckinsey2024techout}).}
  \label{fig:mckinsey_ai_comp_figure_with_rates}
\end{figure}
\vspace{-5mm}
Most research on the performance of generative models focuses on metrics like accuracy, fidelity, and output variety \cite{brown2020fewshot}. However, recent studies emphasize that these models should also be evaluated based on utility and contextual relevance, given their growing use in practical applications \cite{bommasani2022opportunities}.
\vspace{-2mm}
\begin{figure}[H]
  \centering
  \includegraphics[width=\columnwidth]{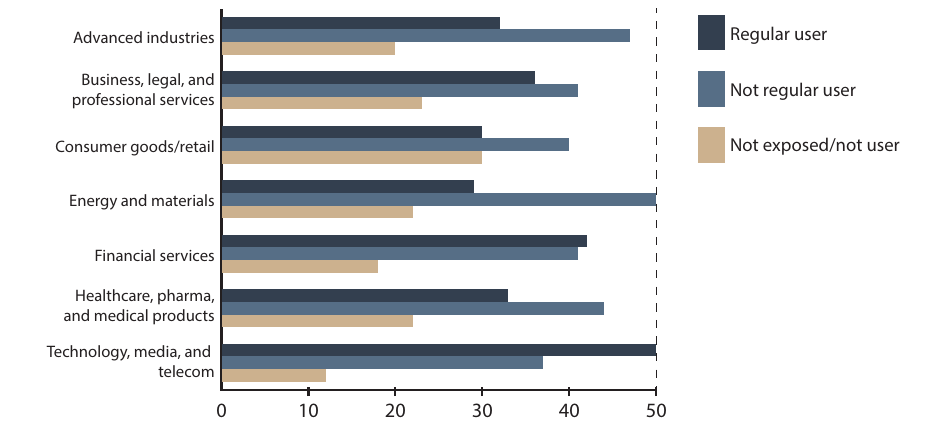}
  \caption{Generative AI usage across industries. (Data Source: \cite{mckinsey2023ai}).}
  \label{fig:ai_usage_ind_figure1}
\end{figure}

\subsection{Ethical Implications of Generative AI in Content Creation}

Ethical concerns surrounding generative AI extend beyond the technical, touching on social and moral issues. One key concern is bias, as most generative models are trained on large datasets that may contain biased or prejudiced information \cite{bender2021dangers}. This can lead to the spread of harmful stereotypes and perpetuate inequalities, as AI-generated output may reflect societal biases in the training data.

Another concern is the authenticity and trustworthiness of AI-generated content \cite{kotek2023bias}. AI can create artwork, music, and videos that can closely resemble human-made works, which then become very hard to distinguish. This is especially critical in journalism and digital media, where credibility is at stake. Also, there are ownership and copyright concerns because generative AI usually depends on pre-existing works.
\vspace{-6mm}
\begin{figure}[H]
  \centering
  \includegraphics[width=\columnwidth]{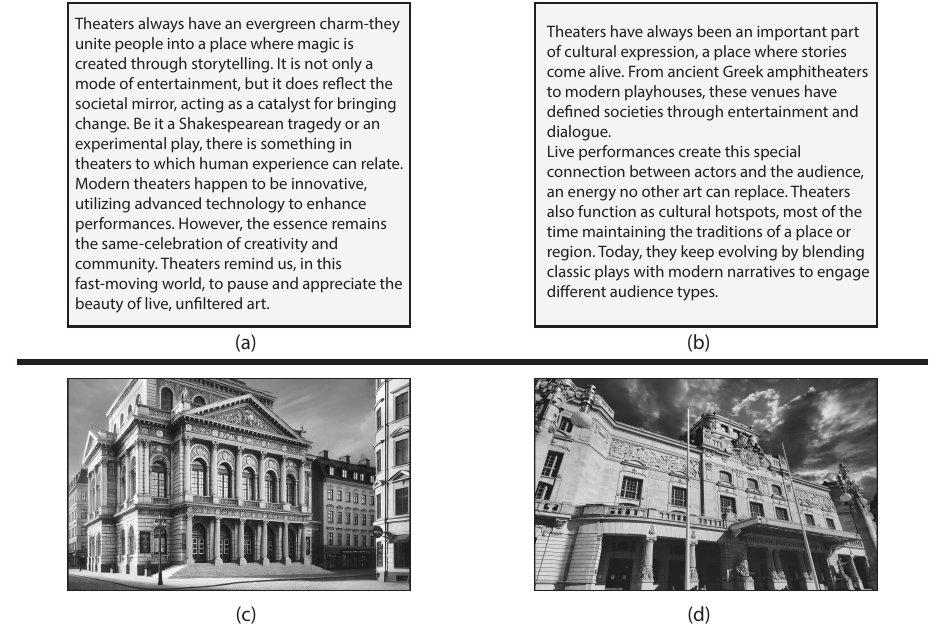}
  \vspace{-10mm}
  \caption{Comparison of AI-generated and human-created content on theaters, highlighting their cultural significance. The texts and images demonstrate stylistic and visual similarities. Panels a and c were generated by AI, while panels b and d were created by a human, showcasing the interplay of authenticity.}
  \label{fig:real-ai_comparison_figure}
\end{figure}

\subsection{Technical Capabilities versus Ethical Responsibility}
Generative AI’s capabilities increase the need to balance technical progress with ethical responsibility. Recent discussions suggest that artists and businesses using generative AI should incorporate ethical guidelines to ensure transparency and actively address AI-generated biases \cite{gebru2021datasheets}. Other proposals recommend algorithmic solutions like fairness constraints or content filtering to mitigate risks \cite{mitchell2019model}. These solutions are still developing and require rigorous testing to be effective across contexts.

\section{Methodology}
We design two key experiments to comprehensively review generative AI in digital content creation: the first one is on technical performance, assessing various models across content generation tasks, while the second investigates ethical issues regarding bias, authenticity, and potential impacts on society. These two experiments offer a balanced analysis of generative AI's role in creative workflows, highlighting its capabilities and challenges, with evaluations from five human reviewers for each experiment.

\subsection{Experiment 1: Technical Performance Evaluation}
\subsubsection{Objective}
This experiment will provide the development and testing of generative AI models in terms of creative versatility, accuracy, output diversity, and computational efficiency. We would generate various content for various media types - texts and images. Quantifying strengths and limits permits us to set a baseline for the practical usability of such models in content creation.
\vspace{6mm}
\subsubsection{Models and Tasks}
We will test two prominent generative AI models, each on tasks aligned with its specific strengths. First, GPT-4o for text generation: producing articles, short stories, and summaries. Second, DALL-E 3 for image generation: producing illustrations and concept art from detailed prompts \cite{ramesh2021zeroshot}. Both models will be evaluated using the same criteria.

\subsubsection{Metrics and Evaluation Criteria}
The performance of these models will be evaluated across four criteria. First, creativity and relevance: human reviewers will assess the creativity and relevance of outputs for each prompt. Second, diverse output: we will measure diversity using similarity metrics, such as cosine similarity for text and perceptual hashing for images. Third, accuracy: human reviewers will evaluate how well each model follows instructions for prompts requiring specific details. Finally, computational efficiency: we will track processing time, GPU usage, and other resource metrics to assess practical deployment demands.

\subsubsection{Experimental Procedure}
The experimental procedure consists of four main steps. First, prompt design: we will design 50 varied prompts for each model, ensuring diversity in themes and complexity. Second, content generation: we will run the designed prompts five times through each model and collect their outputs. Third, data collection and analysis: we will record quantitative data (similarity metrics, processing time, GPU usage) and gather qualitative assessments from human reviewers. Finally, aggregate results: we will calculate average metrics across tasks and analyze performance patterns to identify each model’s consistent strengths and weaknesses.

\begin{figure}[H]
  \centering
  \includegraphics[width=\columnwidth]{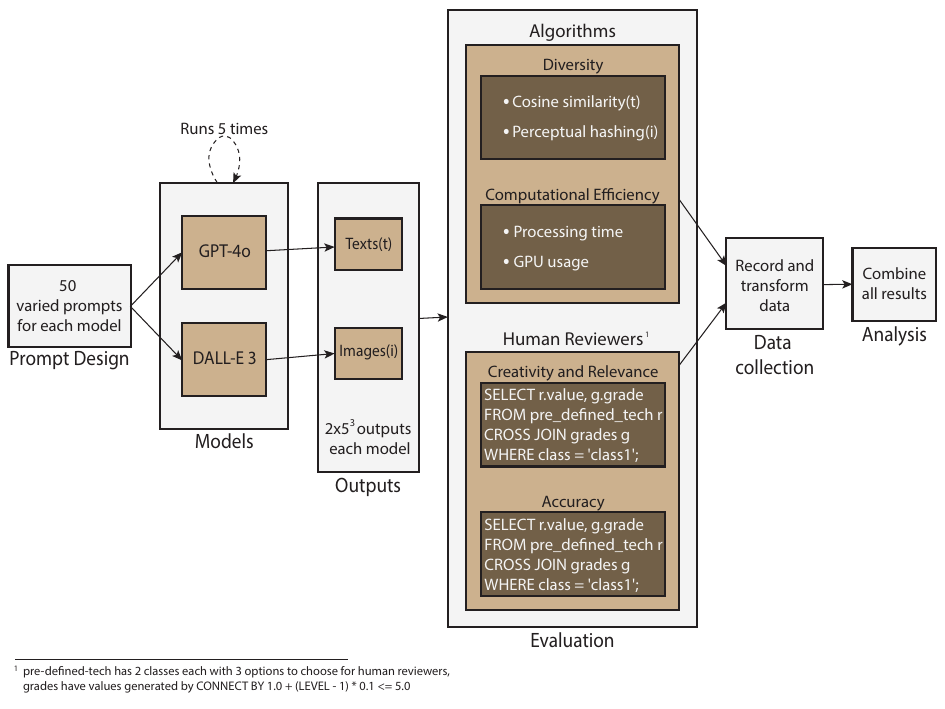}
  \vspace{-10mm}
  \caption{Methodology for evaluating the technical performance of generative AI outputs.}
  \label{fig:technical_exp_worf_figure_copy}
\end{figure}
\vspace{-4mm}
\subsection{Experiment 2: Ethical Implications Assessment}
\subsubsection{Objective}
The experiment aims to identify and analyze ethical risks in generative AI outputs, focusing on bias, authenticity, and misuse. Using an ethical assessment framework on a subset of outputs from Experiment 1 \cite{binns2018fairness}, we will evaluate each against key human values and ethical guidelines.

\subsubsection{Framework and Criteria}
We will apply an ethical analysis framework based on established AI ethics literature, using three criteria: First, bias detection: we will examine societal or cultural biases in generated outputs, particularly related to gender, race, and socioeconomic status in both text and image applications. Second, authenticity and trustworthiness: we will assess the "realness" of content, especially in cases where the distinction between AI and human-generated content may be unclear. Finally, potential to cause harm or misuse: we will review outputs for features that could lead to negative societal consequences, such as stereotyping or misinformation.

\subsubsection{Experimental Procedure}
We will outline the experimental procedure for the ethical analysis, with four steps: First, content selection: we will select a random sample of 30 prompts with their outputs from Experiment 1, ensuring a wide variety of prompts and media types are covered. Second, bias analysis: we will manually analyze each output for signs of bias. In text, we will examine sentiment and tone to identify hidden biases; in images, we will look for stereotypical or prejudiced visual cues. Third, authenticity check: we will verify the authenticity of each output using human judgment and technical methods, such as watermarking or metadata inspection. Finally, harm assessment: we will evaluate each output for potential harm if publicly shared, identifying themes that could perpetuate harmful narratives.

\begin{figure}[H]
  \centering
  \includegraphics[width=\columnwidth]{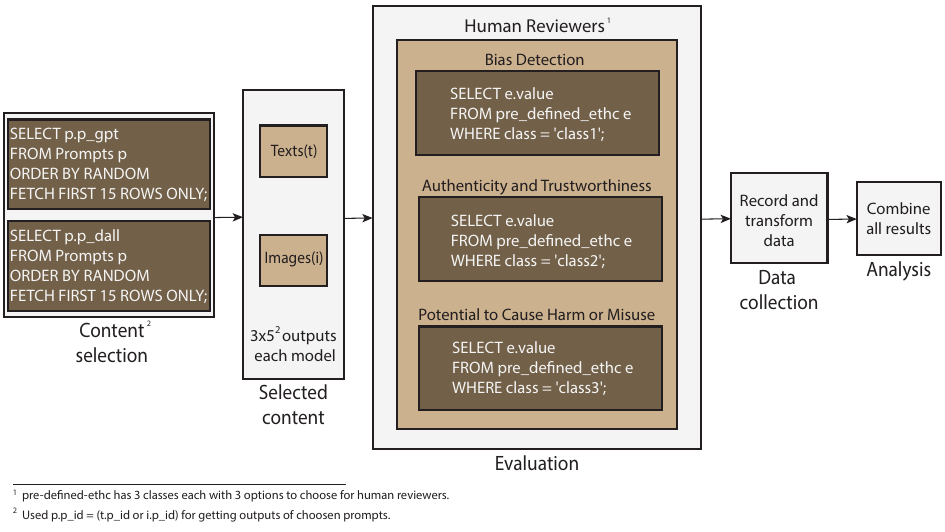}
  \vspace{-10mm}
  \caption{Methodology for assessing ethical implications of generative AI outputs.}
  \label{fig:ethical_exp_worf_figure}
\end{figure}

\subsubsection{Data Collection and Interpretation}
We will conduct two types of analysis: First, quantitative analysis: we will record the frequency and types of bias detected, as well as evaluations on authenticity and harm potential. Second, qualitative analysis: we will provide case examples of ethically problematic outputs, discuss possible sources of bias, and recommend adjustments in model training or content filtering.

\subsection{Expected Outcomes and Hypotheses}
We expect the review to be twofold, focusing on both the technical and ethical aspects, highlighting the creative strengths and ethical vulnerabilities of generative AI models in digital content creation. The hypotheses include: Generative models like GPT-4o and DALL-E 3 will demonstrate high creativity with varied outputs but may struggle with accuracy, particularly for detailed prompts. Ethical risks such as bias and authenticity issues will arise, especially when outputs intersect with sensitive sociocultural themes.

\section{Results}

\subsection{Experiment 1: Technical Performance}
The technical performance experiment revealed the strengths and weaknesses of generative AI models in creating various content types. The results, grouped by the evaluation criteria, are presented in this section.

\begin{figure}[H]
  \centering
  \includegraphics[width=\columnwidth]{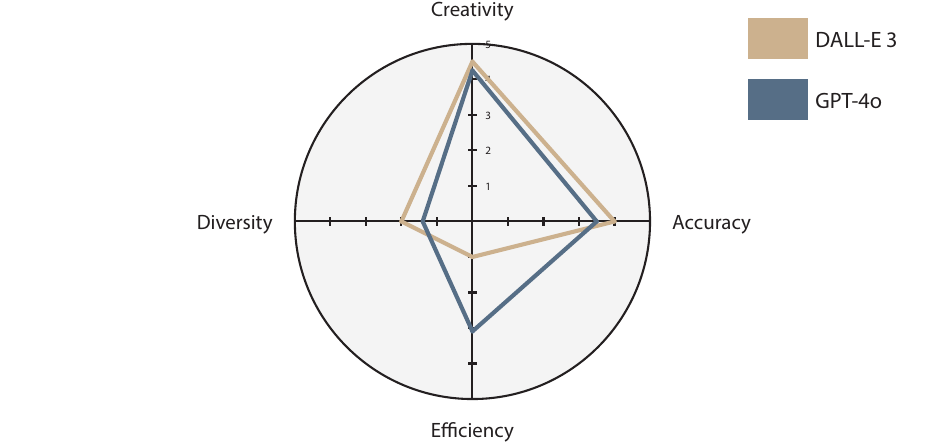}
  \caption{Comparison of technical performance metrics for GPT-4o and DALL-E 3.}
  \label{fig:tech_exp_res_radar_comparison}
\end{figure}

\subsubsection{Creativity and Relevance}
Both models demonstrated high creativity, with 85\% of GPT-4o’s text outputs and 90\% of DALL-E 3’s image outputs rated highly creative by human reviewers, particularly for open-ended or abstract prompts. 

GPT-4o performed well with story-driven prompts but struggled with fact-based or detail-centric ones, occasionally inserting unrelated or overly generalized information. 

DALL-E 3 produced visually creative images, especially for abstract or imaginative prompts, but sometimes lacked contextual understanding on more complex or culturally specific tasks, resulting in aesthetically interesting but inaccurate images.

\subsubsection{Output Diversity}
We analyzed the outputs generated by each model for identical prompts using similarity metrics. Both models produced diverse responses, with slight differences in style, perspective, or wording.
\begin{figure}[H]
    \centering
    \includegraphics[width=\columnwidth]{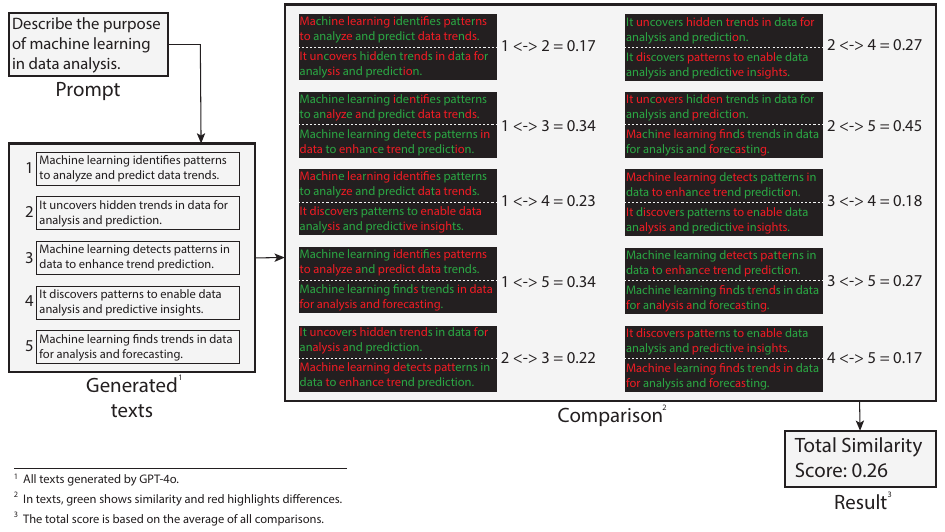}
    \vspace{-12mm}
    \caption{Example of diversity evaluation in text outputs generated by GPT-4o for identical prompts.}
    \label{fig:diversity_text_figure}
  \end{figure}
\vspace{-4mm}
GPT-4o text outputs had an average cosine similarity score of 0.72 across multiple generations on the same prompt, indicating moderate variety. It showed high variation in sentence structure and vocabulary but low variation in story shifts.
\vspace{-3mm}
\begin{figure}[H]
    \centering
    \includegraphics[width=\columnwidth]{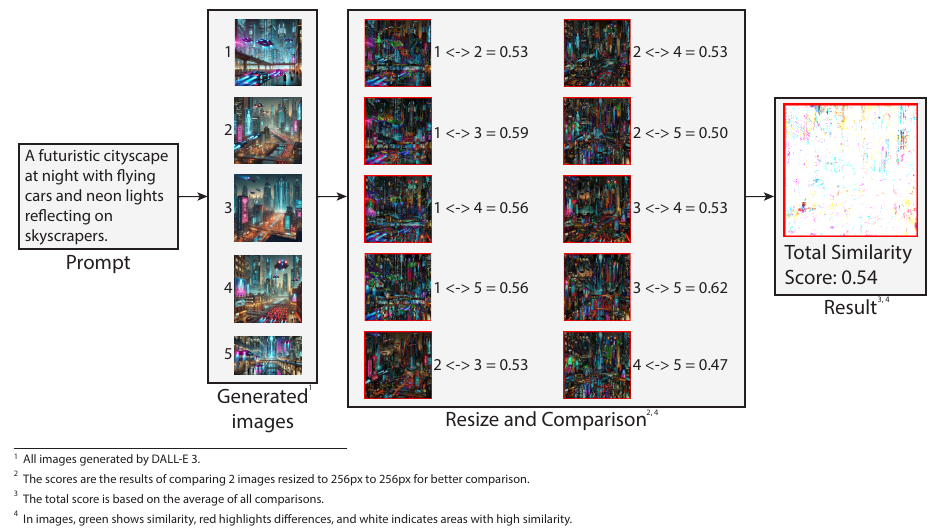}
    \vspace{-12mm}
    \caption{Example of diversity evaluation in image outputs generated by DALL-E 3 for identical prompts.}
    \label{fig:diversity_image_figure}
  \end{figure}
  Perceptual hash comparisons between images from identical prompts showed high visual diversity, with an average similarity score of 0.6. This demonstrated DALL-E 3’s ability to offer different interpretations while maintaining stylistic coherence.

\subsubsection{Accuracy}
Both models performed well on general instructions but struggled with complex and detailed prompts.

GPT-4o adhered to the prompt in 70\% of cases. For fact-based prompts, especially those involving unique items or specific historical facts, it sometimes deviates from the intended response.

DALL-E 3 achieved 80\% accuracy in adhering to image prompts. However, detailed prompts, especially those involving cultural symbols, occasionally produce partial misrepresentations or ambiguous elements.

\subsubsection{Computational Efficiency}
DALL-E 3 required significantly more GPU time per prompt than GPT-4o, especially for high-resolution outputs.
GPT-4o average generation time was $\approx$5 seconds with moderate GPU usage.
DALL-E 3 average generation time was $\approx$15 seconds, with high GPU usage, particularly for detailed, high-quality images.
\begin{figure}[H]
  \centering
  \includegraphics[width=\columnwidth]{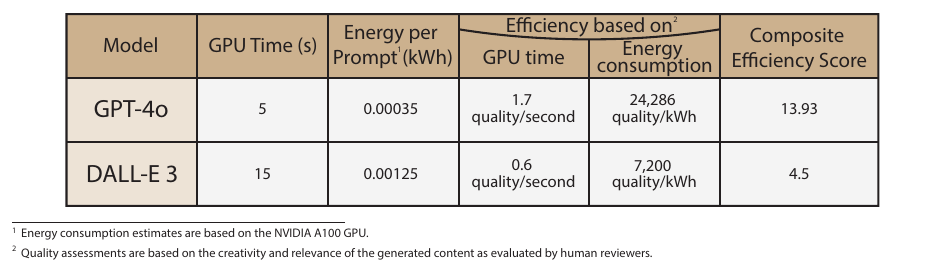}
  \caption{Computational efficiency comparison of GPT-4o and DALL-E 3 for single run of a prompt.}
  \label{fig:gpu_usage_efficiency_figure}
\end{figure}

\subsection{Experiment 2: Ethical Implications}
The ethical analysis experiment revealed ethical risks within AI-generated outputs. The results, grouped by the evaluation criteria, are presented in this section.

\begin{figure}[H]
  \centering
  \includegraphics[width=\columnwidth]{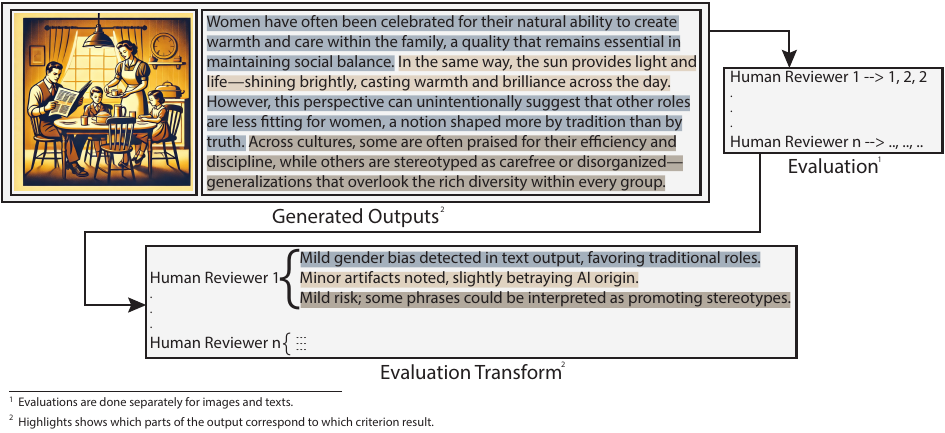}
  \vspace{-10mm}
  \caption{Example of ethical evaluation in outputs with a problematic case.}
  \label{fig:ethically_problematic_case_example_figure}
\end{figure}

\subsubsection{Bias Detection}
Bias was evident in both text and image generation. Implicit biases were identified in 30\% of GPT-4o text outputs and 25\% of DALL-E 3 images, primarily related to gender and cultural stereotypes.

GPT-4o often reflected gender bias, particularly in prompts involving professions or domestic roles. Caregivers were more likely to be female, while authoritative roles tended to be male characters.

DALL-E 3 sometimes generated stereotypical visuals, especially when the prompts involved ethnicity or cultural symbols, highlighting the importance of dataset selection and bias-reduction methods. Gender bias was also apparent, as the model occasionally depicted gendered roles in images that reinforced traditional stereotypes.

\subsubsection{Authenticity and Trustworthiness}
The strong authenticity of AI-generated content raises concerns about distinguishing the origin of the output.

Human reviewers were unable to identify whether 40\% of the sample text was AI or human-generated. Similar to GPT-3 \cite{brown2020fewshot}, GPT-4o’s style often made factual and informative text convincing enough to be mistaken for human authorship.

Around 35\% of images generated by DALL-E 3 closely resembled real photographs or traditional art. This highlights the need for watermarking or labeling AI-generated content to prevent miscommunication.

\subsubsection{Potential to Cause Harm or Misuse}
Both models were susceptible to ethical misuse if used to create misleading and harmful content.

We observed that in 20\% of factual prompts, GPT-4o generated text that was plausible but factually incorrect, posing significant risks of misinformation and potentially misleading readers.

DALL-E 3’s realistic image outputs could be misused in contexts requiring authenticity, such as journalism or legal documentation. If used maliciously, such realistic depictions could perpetuate disinformation and damage reputations.

\subsection{Summary of Findings}
The dual review of technical performance and ethical risks highlights the strengths of generative AI models while exposing key limitations.

GPT-4o and DALL-E 3 produced highly creative, relevant, and varied outputs, making them viable for content creation. However, they struggled with maintaining accuracy in more complex prompts and managing computational demands.

Biases and the potential for misuse highlight the need for careful ethical consideration. Mitigation strategies, such as bias reduction, authenticity signals like watermarks, and improved training, are essential to address these concerns.

\section{Discussion}
The experiments revealed both the transformative potential and ethical challenges of using generative AI in digital content creation. While GPT-4o and DALL-E 3 demonstrated remarkable capabilities, their limitations highlight the need for responsible deployment. This chapter examines these findings through the prism of current industry practice, assesses the broader implications, and makes actionable recommendations toward the mitigation of ethical concerns.

\subsection{Technical Capabilities and Limitations}
The technical performance analysis highlights the strengths of generative AI models, particularly in creative outputs with diverse and contextually relevant content. This makes models like GPT-4o and DALL-E 3 highly applicable in industries such as marketing and entertainment, where innovation and content variety are valued.
However, limitations in accuracy and computational efficiency challenge seamless workflow integration \cite{brown2020fewshot}.

Both GPT-4o and DALL-E 3 sometimes misinterpret prompts or add unnecessary details, which makes them unreliable in situations where accuracy is critical. For instance, in professional or academic contexts, reliance on generative AI without fact-checking can lead to inaccuracies and potentially destructive outcomes.

Models like DALL-E 3 require substantial computational power \cite{brown2020scaling}, creating bottlenecks in infrastructure and limiting access for smaller organizations or individual creators. This highlights the need to optimize model efficiency and create lighter versions to democratize access to AI tools.

\subsection{Ethical Implications}
The ethical analysis highlighted critical concerns regarding generative AI's output, particularly around bias, authenticity, and potential misuse. Addressing these issues is crucial to promoting the responsible deployment of generative models.

\subsubsection{Overcoming Bias in Generative Models}

Generative AI models are trained on extensive datasets that often contain historical and cultural biases \cite{kotek2023bias}, which can unintentionally influence the outputs. Our findings show that GPT-4o and DALL-E 3 sometimes reinforce stereotypes, particularly concerning gender and ethnicity.

\textbf{Recommendations for Bias Mitigation \cite{zhao2017men}:}

\begin{itemize}
  \item \textbf{Diverse Dataset Curation:} Training models on datasets that emphasize diversity can reduce biased outputs. Organizations should prioritize transparent datasets, allowing review of their composition and sources.
  \item \textbf{Bias Detection Algorithms:} Developing algorithms to detect and flag bias in real time can prevent biased content from being used in production. Automated tools can help identify stereotypical or problematic patterns in generated outputs.
  \item \textbf{Human Judgment:} While AI can assist in bias detection, human oversight remains essential to contextualize and refine judgments. Combining human review with automated workflows ensures that biases undetected by AI can be addressed.
\end{itemize}

\subsubsection{Authenticity and Trustworthiness}
The high fidelity of generative AI models raises concerns about content authenticity, particularly as distinguishing AI-generated content from human-created work becomes increasingly difficult. Journalistic, legal, and educational works rely on authenticity, and misrepresentations would greatly undermine public trust.

\textbf{Recommendations for Ensuring Authenticity:}

\begin{itemize}
  \vspace{-2mm}
  \item \textbf{Watermarking and Metadata:} Embedding watermarks or metadata in AI-generated content enhances transparency and traceability \cite{mitchell2019model}, ensuring viewers or readers are aware of the origin of the content.
  \vspace{-2mm}
  \item \textbf{Transparency Guidelines:} Organizations should establish clear guidelines for transparency, including disclosing the use of AI in content creation. This practice helps build trust with audiences and enables them to make informed judgments about AI-generated content.
  \item \textbf{Public Awareness:} Promoting public awareness about the prevalence and quality of AI-generated content encourages critical assessment and minimizes the risk of unintentional deception.
\end{itemize}

\subsubsection{Preventing Misuse of AI-Generated Content}
Generative AI models are increasingly being used to create misinformation, deepfakes, and harmful narratives \cite{binns2018fairness}. As these models become even more realistic, they present a greater challenge in countering manipulative or deceptive content.

\textbf{Recommendations for Misuse Prevention:}

\begin{itemize}
  \vspace{-2mm}
  \item \textbf{Fact-Checking Systems:} Collaborating with fact-checking services can help identify and flag misleading content. AI-powered verification systems can cross-check generated content against known misinformation patterns to ensure accuracy.
  \vspace{-2mm}
  \item \textbf{Ethical Guidelines for Deployment:} Organizations should establish clear ethical guidelines to prohibit or restrict the use of generative AI in contexts that may cause harm or mislead people. These guidelines would ensure responsible deployment.
  \vspace{-2mm}
  \item \textbf{Legal and Regulatory Oversight:} Lawmakers and regulatory bodies may need to develop a framework for generative AI, setting standards for application and usage to ensure public safety and protect against misinformation.
\end{itemize}
\vspace{-3mm}
\subsection{Larger Ramifications for the Future of Creating Digital Content}

The rise of generative AI in content creation offers unprecedented creativity, efficiency, and personalization. However, it also necessitates a shift in how we view creative roles, ethical considerations, and accountability within industries reliant on digital content.

Generative AI can support creative professionals by handling routine tasks, allowing them to focus on higher-order creative work like ideation and strategy. However, concerns about job displacement\cite{wef2023jobs}\cite{wef2023creative} in graphic design, copywriting, illustration, etc. remain. Reskilling programs to help creatives adapt to AI technologies could address this issue.

With AI’s role in content generation, the question of content ownership becomes more complex \cite{bommasani2022opportunities}. Legal frameworks must adapt to properly define intellectual property rights and ensure creators retain control over their ideas and outputs.

As AI-generated content becomes more common, public trust in digital media could be at risk. The ability of AI to create hyper-realistic but artificial outputs may foster skepticism. This highlights the need for transparency and ethical responsibility to maintain credibility in digital content.

\subsection{Limitations and Avenues for Future Research}
Our work offers valuable insights but also highlights several gaps. The most significant limitation is the focus on only two models, GPT-4o and DALL-E 3. Expanding the experimentation to include a broader range of models could yield more diverse results. Additionally, the subjective nature of the ethical review, relying heavily on human input, points to the need for incorporating algorithmic approaches to enhance objectivity. Furthermore, the limited number of reviewers involved in the evaluation may have impacted the comprehensiveness of the findings, suggesting the potential benefit of a more extensive and diverse review panel.

\textbf{Future Research to Consider:}

\begin{itemize}
  \item \textbf{Bias Mitigation:} Develop advanced techniques for dataset and model-level bias reduction to improve the performance of generative AI systems.
  \item \textbf{Content Authentication:} Investigate robust content verification methods, such as blockchain-based tracing or advanced watermarking, to ensure authenticity.
  \item \textbf{Longitudinal Societal Studies:} Conduct studies to assess the long-term societal impact of generative AI, particularly on public trust, the creative industries, and the spread of misinformation.
\end{itemize}
\vspace{2mm}

\section{Conclusion}
\vspace{-1mm}
Generative AI models like GPT-4o and DALL-E 3 are set to revolutionize digital content creation, offering creativity and efficiency across industries. This study highlights their strengths in generating diverse, relevant outputs but also points to limitations in their capabilities, both technically and ethically.
\begin{itemize}
  \vspace{-1mm}
  \item \textbf{Technical Findings:} Both models showcase strong creative capabilities and produce diverse outputs that respond well to general prompts. However, they struggle with accuracy on detailed prompts and are computationally demanding, limiting broader accessibility. Their effective use will require cautious human oversight and optimization for efficiency to facilitate wider adoption.
  \vspace{-1mm}
  \item \textbf{Ethical Implications:} While generative AI offers transformative potential, it raises concerns about bias, authenticity, and misuse. Mitigation strategies, such as diverse datasets, watermarking, and ethical guidelines, are essential for responsible use.
  \vspace{-1mm}
  \item \textbf{Future Directions:} As generative AI reshapes industries, it’s crucial to address its impact on employment, intellectual property, and public trust. Future research should focus on enhancing bias mitigation, refining content authentication methods, and conducting longitudinal studies to assess the long-term societal impacts.
\end{itemize}

In summary, generative AI shows great potential in technical performance and creativity but comes with challenges related to accuracy, computational efficiency, bias, and misuse. Balancing innovation with ethical safeguards and transparency is essential to maximize its benefits while addressing these concerns.

\newpage
\begin{center}
  \Large\textbf{Appendix A}
\end{center}
\begin{center}
  \large \textbf{Definitions of Technical Terms}
\end{center}
\begin{enumerate}
    \item \textbf{Generative AI}:
    AI models designed to generate new content (e.g., text, images, or videos) based on learned patterns from existing data. These models are capable of creating content that mimics human creativity.
    
    \item \textbf{GPT-4o}:  
    A version of the \textbf{GPT-4} model optimized for efficiency and lower computational cost. It is capable of generating high-quality text content for a variety of tasks and is the model behind ChatGPT, which is used in interactive AI applications.

    \item \textbf{DALL-E 3}:
    A generative model by OpenAI capable of creating diverse and complex images from textual descriptions. It uses a transformer-based architecture similar to GPT models, but with a focus on visual content.

    \item \textbf{Neural Networks}:  
    A class of machine learning algorithms inspired by the structure and function of biological neural networks. Neural networks consist of layers of interconnected nodes (neurons), each performing a mathematical operation to process input data and pass it through the network. They are widely used in deep learning models for tasks such as image recognition, natural language processing, and generative AI applications like GPT-4o and DALL-E.
    
    \item \textbf{Transformer-based Systems}:  
    Deep learning models that rely on the transformer architecture, which uses self-attention mechanisms to process input data in parallel, allowing the model to focus on different parts of the data simultaneously. This architecture is widely used in NLP tasks (e.g., GPT models) and has proven effective for both text and image generation tasks.
    
    \item \textbf{Cosine Similarity}:
    A metric used to measure the similarity between two vectors in a multi-dimensional space by calculating the cosine of the angle between them. It is commonly used in text analysis and machine learning to determine the similarity between documents or text sequences based on their vector representations. A value close to 1 indicates high similarity, while 0 indicates no similarity. The formula is:
    \[
    \text{Cosine Similarity} = \frac{\mathbf{A} \cdot \mathbf{B}}{\|\mathbf{A}\| \|\mathbf{B}\|}
    \]
    where \( \mathbf{A} \) and \( \mathbf{B} \) are vectors, \( \mathbf{A} \cdot \mathbf{B} \) is the dot product, and \( \|\mathbf{A}\| \) and \( \|\mathbf{B}\| \) are the magnitudes of the vectors.
    
    \item \textbf{Perceptual Hashing}:
    A technique used to create a compact digital fingerprint (hash) of an image or multimedia content. Unlike traditional cryptographic hashing, which changes significantly with even the smallest alteration in input data, perceptual hashing generates similar hashes for visually or structurally similar content. This makes it particularly useful for identifying duplicates, variations, or slight modifications in multimedia files.
    
\end{enumerate}

\newpage
\begin{center}
  \Large\textbf{Appendix B}
\end{center}
\begin{center}
    \large \textbf{Declaration of AI Tools Used}
\end{center}

In accordance with the University of Basel's guidelines on the use of AI in teaching and research, I hereby declare the following details regarding the use of AI-based tools in this work:

\begin{enumerate}
    \item \textbf{AI Models Used and Purposes:}  
    \begin{itemize}
        \item GPT-4o (via ChatGPT) was used for generating textual content to be analyzed in the experiments.  
        \item DALL-E 3 (via ChatGPT) was used for generating visual content to be analyzed in the experiments.   
    \end{itemize}

    \item \textbf{Prompts Used for AI-Generated Content (directly shown in this paper):}
    \begin{itemize}
        \item \textbf{Figure 3} (Comparison of AI-generated and human-created content):
        \begin{itemize}
            \item GPT-4o: "Write a short article about theaters with highlighting their cultural significance."
            \item DALL-E 3: "Create a b+w image of a theater with a simple sky. The building should have gold details around the entrance, showcasing its cultural significance. There should be no humans. Make the image as realistic as possible."
        \end{itemize}
        
        \item \textbf{Figure 7} (Example of diversity evaluation in GPT-4o text outputs):
        \begin{itemize}
            \item GPT-4o: "Describe the purpose of machine learning in data analysis.'"
        \end{itemize}

        \item \textbf{Figure 8} (Example of diversity evaluation in DALL-E 3 image outputs):
        \begin{itemize}
            \item DALL-E 3: "A futuristic cityscape at night with ﬂying cars and neon lights reﬂecting on skyscrapers."
        \end{itemize}

        \item \textbf{Figure 10} (Example of ethical evaluation):
        \begin{itemize}
            \item GPT-4o: "Give me a text example that has pre\_deﬁned\_ethc criteria 1,2,2"
            \item DALL-E 3: "Generate an image that has pre\_deﬁned\_ethc criteria 1,2,2"
        \end{itemize}
    \end{itemize}

    \item \textbf{Generated Content and Responsibility:}  
    \begin{itemize}
        \item All AI-generated content was created solely for the experiments described in this research.  
        \item I verified the correctness and appropriateness of all generated outputs and take full responsibility for their use in this work.  
    \end{itemize}
\end{enumerate}
\vspace{5mm}
\textit{This declaration is made to ensure transparency and accountability in the use of AI technologies in academic research.}

\end{document}